\documentclass[11pt,a4paper,logo,copyright,nonumbering]{xiaomi}

\usepackage[authoryear,sort&compress,round]{natbib}
\usepackage{etoolbox}

\usepackage{amsmath,amsfonts,bm}

\def\eqref#1{equation~\ref{#1}}
\def\Eqref#1{Equation~\ref{#1}}

\def\1{\bm{1}}

\DeclareMathAlphabet{\mathsfit}{\encodingdefault}{\sfdefault}{m}{sl}
\SetMathAlphabet{\mathsfit}{bold}{\encodingdefault}{\sfdefault}{bx}{n}

\definecolor{xiaomiorange}{HTML}{FF6901}
\colorlet{tableheaderbg}{xiaomiorange!10}
\newcommand{\tableheader}[1]{\textcolor{black!85}{\textbf{#1}}}
\newcommand{\ourtool}{\textsc{Gagar}}
\newcommand{\flash}{Flash}
\newcommand{\pro}{Pro}
\newcommand{\mimoflash}{MiMo-V2.6-Flash}
\newcommand{\mimopro}{MiMo-V2.6-Pro}

\title{\centering Groupwise Agentic Grading and Advantage Redistribution for Code Agent RL}
\newcommand{\paperauthors}{\parbox[t]{\textwidth}{\centering\normalfont\sffamily\fontsize{11}{15}\selectfont
    Jinhao Dong\textsuperscript{1,2}\quad
    Liang Zhao\textsuperscript{1}\quad
    Zihao Yue\textsuperscript{1,2}\quad
    Wenhan Ma\textsuperscript{1,3}\quad
    Linghao Zhang\textsuperscript{1}\\[3pt]
    Lei Li\textsuperscript{1,4}\quad
    Shicheng Li\textsuperscript{1}\quad
    Yifan Song\textsuperscript{1}\quad
    Bowen Ye\textsuperscript{1,3}\quad
    Fuli Luo\textsuperscript{1,\textdagger}\\[7pt]
    \textsuperscript{1}LLM Core, Xiaomi\qquad
    \textsuperscript{2}Renmin University of China\\[2pt]
    \textsuperscript{3}Peking University\qquad
    \textsuperscript{4}University of Hong Kong\\[4pt]
    {\fontsize{9}{11}\selectfont\textsuperscript{\textdagger}Corresponding author.}
  }}
\hypersetup{pdfauthor={Jinhao Dong, Liang Zhao, Zihao Yue, Wenhan Ma, Linghao Zhang, Lei Li, Shicheng Li, Yifan Song, Bowen Ye, Fuli Luo}}

\author{\protect\paperauthors}
\hypersetup{
    pdftitle={Groupwise Agentic Grading and Advantage Redistribution for Code Agent RL},
    colorlinks=true,
    linkcolor=black,
    citecolor=black,
    urlcolor=black
}

\begin{abstract}
Reinforcement learning (RL) for code agents often uses executable tests to provide binary rewards. With these rewards, Group Relative Policy Optimization (GRPO) assigns identical advantages to test-passing trajectories within each rollout group, overlooking differences in implementation quality and adherence to task requirements. This leaves the policy without a learning signal that favors clean, targeted implementations over those containing unnecessary or out-of-scope changes. We introduce \ourtool{}, a framework for quality-aware credit redistribution in code agent RL. Built on dynamic sampling that retains groups containing both passing and failing trajectories, \ourtool{} places all trajectories from each group in a shared workspace, where an SFT-trained agentic grader jointly inspects them and ranks the test-passing candidates. Based on this ranking, we downweight lower-ranked trajectories and proportionally rescale the advantages of all test-passing trajectories to restore their original sum. This sum-preserving redistribution retains the relative weights established by quality-based downweighting while shifting credit toward higher-quality implementations. We evaluate \ourtool{} at industrial scale using pre-RL SFT checkpoints of MiMo-V2.6-Flash (310B total parameters) and MiMo-V2.6-Pro (1.02T total parameters). Controlled code-only \flash{} experiments show improved code agent performance, reduced trajectory-length growth, and more stable training. We further apply \ourtool{} in large-scale mixed-task RL with both \flash{} and \pro{}. Our results support combining test-based verification with groupwise agentic grading to improve the quality and stability of code agent RL.

\end{abstract}

\begin{document}

\begingroup
\patchcmd{\maketitle}{\vskip20pt}{\vskip10pt}{}{}
\patchcmd{\maketitle}{\vskip30pt}{\vskip12pt}{}{}
\patchcmd{\abscontent}{\vspace{5ex}}{\vspace{2ex}}{}{}
\maketitle
\endgroup

\section{Introduction}
\label{sec:introduction}
Reinforcement learning (RL) with executable feedback provides a scalable approach to training code agents that inspect repositories, modify code, and validate their solutions over long interactions.
In a common setup, each trajectory receives a binary reward according to whether its submitted implementation passes the tests.

With group-relative advantage estimation, as used in GRPO~\citep{shao2024deepseekmath}, test-passing trajectories within the same rollout group receive identical outcome advantages. This provides a clear signal for learning to solve a task, but leaves an important question unanswered: \emph{among successful trajectories, which ones deserve stronger reinforcement?}

Passing the same tests does not imply equal implementation quality or equal practical value to developers. One solution may address the root cause with a focused change that follows repository conventions, while another introduces unnecessary complexity, weakens validation, or modifies code outside the requested scope. These differences shape the developer experience: they affect how much review and rework are needed before a patch can be accepted and merged, and how easily the resulting code can be maintained. Successful trajectories also differ in how effectively they gather evidence, identify a solution, and validate their changes. Binary feedback overlooks these distinctions when the test outcomes are identical. Our objective is to reinforce sound, efficient problem-solving strategies and precise, minimally invasive implementations that fully satisfy task requirements. In doing so, we aim to produce maintainable, merge-ready code and provide a reliable, low-friction user experience.

Turning these distinctions into useful training signals requires both reliable assessment and appropriate credit allocation. Comparing successful trajectories from the same task can reveal unnecessary complexity or ineffective strategies that are difficult to recognize in isolation. We therefore adopt groupwise assessment rather than scoring trajectories independently. Reliable comparison also requires repository context and execution evidence that static, chat-based assessments may miss, motivating an agentic evaluator that can inspect code and run checks. Finally, simply downweighting lower-quality passes reduces the group's total positive advantage while leaving negative advantages unchanged. We therefore seek to redistribute credit among successful trajectories rather than merely weaken their overall positive training signal.

We introduce \ourtool{} (\textbf{Groupwise Agentic Grading for Advantage Redistribution}), a quality-aware framework for code agent RL. For each mixed-outcome rollout group, an agentic grader receives a shared workspace containing the task specification, repository, submitted patches, execution results, and trajectories. It can inspect code and run targeted checks before ranking the test-passing candidates, allowing ties when the evidence is inconclusive. The grader compares different implementations for the same task to identify unnecessary changes and potential problems.
The resulting ranking reflects relative quality within the group and guides credit redistribution from lower-quality to higher-quality implementations.

The grader ranks passing solutions along five dimensions: suitability of the solution approach, implementation precision, minimality of changes, avoidance of unintended side effects, and consistency with codebase conventions. \ourtool{} uses these rankings to downweight lower-quality passing trajectories, then proportionally rescales the advantages of all passing trajectories. In the sum-preserving formulation, this restores their original total positive advantage while retaining the ranking-based relative weights and leaving failed-trajectory advantages unchanged.

Our primary controlled study evaluates \ourtool{} in code-only RL initialized from a pre-RL SFT checkpoint of MiMo-V2.6-Flash (310B total, 15B active parameters) on DeepSWE v1.1~\citep{huang2026deepswe} and SWE-bench Pro. Compared with binary-reward training, \ourtool{} improves later-stage DeepSWE pass rates and training stability while reducing interaction turns and token usage on both benchmarks. We further assess implementation quality and problem-solving behavior through a blinded rubric-based evaluation, reporting average win rates among test-passing solutions. We also examine how sum-preserving redistribution relates to credit balance and training stability. We further apply \ourtool{} in large-scale mixed-task RL with both MiMo-V2.6-Flash and MiMo-V2.6-Pro (1.02T total, 42B active parameters), starting from their respective pre-RL SFT checkpoints. After mixed-task RL, the DeepSWE v1.1 \texttt{avg@3} scores reach 67.9 and 71.9 for \flash{} and \pro{}, respectively (Section~\ref{sec:exp_mixed_rl}).

In summary, our contributions are:
\begingroup
\setlength{\leftmargini}{1.5em}
\begin{itemize}
    \item A grading framework that combines groupwise grading with agentic repository inspection and execution checks to identify quality differences that isolated, text-only assessments may miss.
    \item A quality-aware, zero-sum advantage redistribution formulation that shifts credit from lower- to higher-quality solutions while preserving total positive advantage and its balance with negative advantages.
    \item Industrial-scale validation from pre-RL SFT checkpoints of MiMo-V2.6-Flash and MiMo-V2.6-Pro, demonstrating improvements in task performance, efficiency, and training stability.
\end{itemize}
\endgroup

\section{Related Work}
\label{sec:related_work}

\paragraph{Reinforcement Learning For LLM Agents.}
Reinforcement learning improves agents' multi-step decision-making through feedback from interactions with external environments.
Agent Lightning decouples agent execution from RL training~\citep{luo2025agent},
while HiPER assigns credit across planning and execution~\citep{peng2026hiper}.
GRPO estimates group-relative advantages~\citep{shao2024deepseekmath},
and DAPO filters uniform-outcome groups through dynamic sampling~\citep{yu2025dapo}.
We build on this setting to distinguish implementation quality among test-passing coding trajectories.

\paragraph{Reward Modeling And Shaping.}
Reward modeling and shaping can enrich training feedback beyond task success by assessing solution quality and problem-solving behavior.
ReCode adds candidate-wise process rewards for passing solutions~\citep{fan2026recode},
and TRIAGE supplies segment-level supervision~\citep{xu2026triage}, whereas we compare complete implementations within a task.
Performance-based rewards also target runtime efficiency~\citep{feng2026towards}.
CPO~\citep{ye2025cpo}
and GRRM~\citep{yang2026grrm} use comparative evaluation for dialogue and translation, respectively.
Agent-as-a-Judge evaluates task artifacts through agentic inspection~\citep{zhuge2025agent}.
Groupwise Ranking Reward ranks verifier-passed multimodal reasoning with a single-pass, text-based grader~\citep{jia2026prioritizing}.
Our comparisons instead use interactive inspection of coding trajectories, patches, and repositories, including targeted execution checks.

\paragraph{Credit Assignment.}
Credit assignment determines how feedback is allocated across the decisions and trajectories that produce an outcome.
Temporal approaches include RUDDER's return decomposition~\citep{jose2019rudder}
and FACTOR's trajectory-to-action and action-to-token allocation~\citep{ma2026factor}.
GiGPO~\citep{feng2025gigpo}
and GraphGPO~\citep{cheng2026beyond} exploit shared intermediate states, which are difficult to align in code agent tasks, where trajectories often span over 100 turns and produce divergent repository and execution states.
We instead redistribute credit across successful implementations without intermediate-state alignment.

At the optimization level, reward and advantage shaping adjust which trajectories or tokens receive stronger reinforcement.
PAPO adds rubric-based advantages normalized within the passing subset, yielding a zero-sum additive correction~\citep{tan2026papo}.
EDAS reshapes failed-trajectory advantages using error diversity~\citep{liu2026edas},
while GTPO/GRPO-S~\citep{tan2025gtpo}
and RL-ZVP~\citep{le2026zvp} use entropy-guided shaping, the latter on zero-variance groups.
We instead reweight successful-trajectory advantages in mixed-outcome groups using groupwise implementation-quality ranks, preserving total positive credit and quality-induced weight ratios.

\section{Approach}
\label{sec:method}

\begin{figure}[t]
    \centering
    \includegraphics[width=\linewidth]{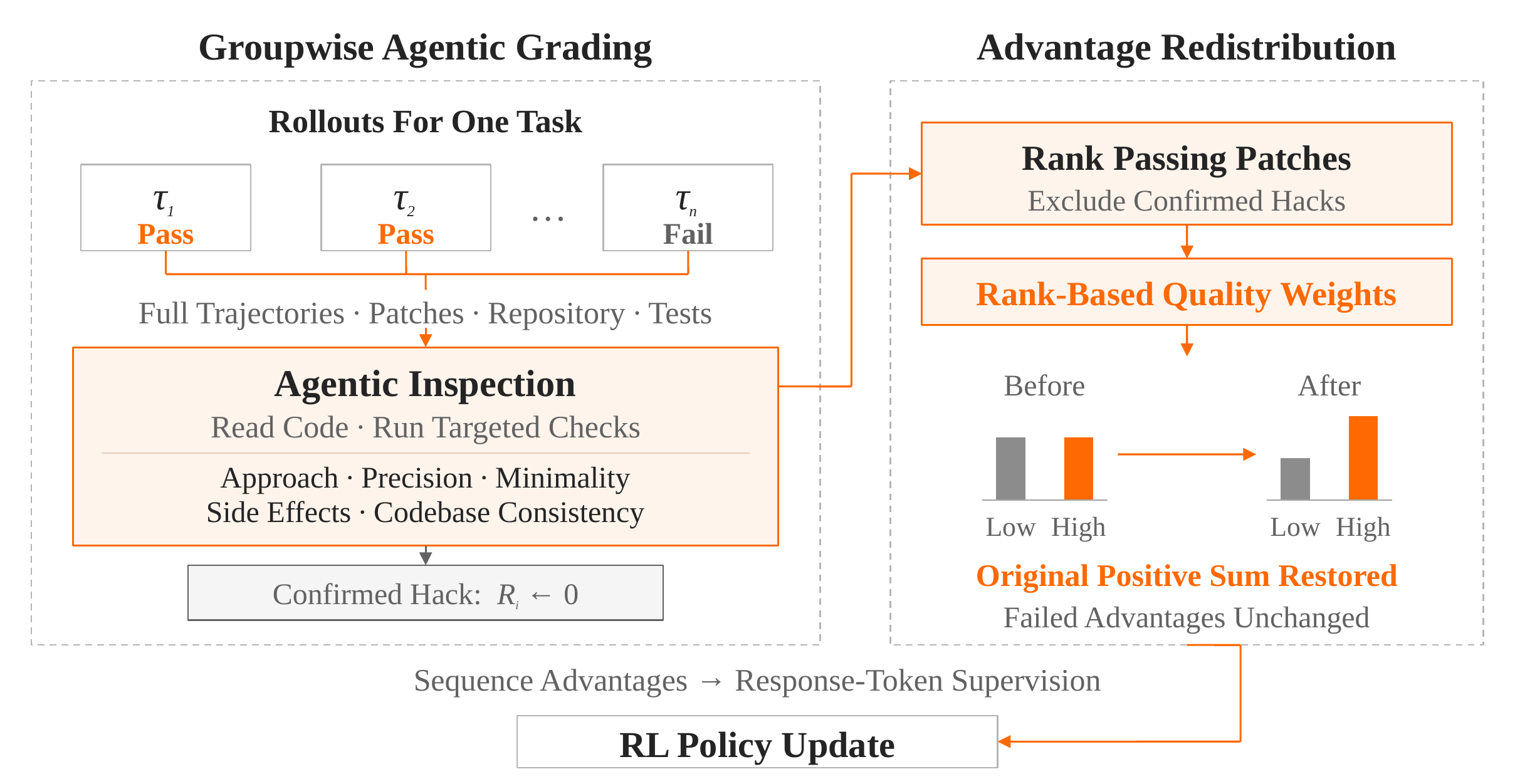}
    \caption{Overview of \ourtool{}. Groupwise agentic grading ranks passing implementations and guides sum-preserving advantage redistribution.}
    \label{fig:credit_architecture}
\end{figure}

\ourtool{} augments reinforcement learning for code agents with groupwise quality supervision beyond binary task outcomes.  Figure~\ref{fig:credit_architecture} shows how groupwise agentic grading is integrated into the RL training loop. For each task, the grader jointly examines successful and failed attempts using their full trajectories, submitted patches, repository context, and test results. It can inspect code and run targeted checks before ranking the valid passing implementations. These rankings determine relative weights on positive advantages; sum-preserving redistribution then shifts credit toward higher-quality solutions while preserving the total positive advantage and leaving failed-trajectory advantages unchanged. The resulting sequence-level advantages supervise the model-generated response tokens during the policy update. We present the core method in this section; Appendix~\ref{sec:method_safeguards} explains how to integrate it with reward adjustments such as length penalties.

\subsection{Training Setup}
\label{sec:method_setup}

Let $x$ denote a coding task specified by its requirements, initial repository state, and execution environment. For each task, the rollout policy generates multiple trajectories, each comprising the agent's responses, tool calls, and environment observations. The final patch produced by each trajectory is evaluated using executable tests, yielding a binary outcome reward.

After excluding trajectories that cannot be evaluated because of infrastructure failures, let $\mathcal{G}_x=\{\tau_i\}_{i=1}^{n}$ denote the group of $n$ valid trajectories. Confirmed hacks receive zero reward and are treated as failures. We denote the resulting effective reward by $R_i\in\{0,1\}$ and compute all group statistics over these $n$ trajectories.
Following Dr.~GRPO~\citep{liu2025doctor}, we use the mean-centered outcome advantage $A_i=R_i-\bar R$, where $\bar R=n^{-1}\sum_{j=1}^{n}R_j$, without normalizing by the group reward standard deviation.

We adopt dynamic sampling~\citep{yu2025dapo} to retain groups containing both successful and failed trajectories, so that $0<\bar R<1$. Let $\mathcal{P}=\{i:R_i=1\}$ and $\mathcal{F}=\{i:R_i=0\}$ denote the passing and failing subsets, respectively. All passing trajectories receive the same positive advantage $1-\bar R$, leaving differences in implementation quality and problem-solving behavior undifferentiated. Our objective is to allocate credit according to these quality differences while preserving the total positive advantage assigned to the group.

\subsection{Groupwise Agentic Grading}
\label{sec:method_judging}

\paragraph{Groupwise Comparison.}
For each task, the grader jointly examines all rollouts in a shared workspace containing the task specification, repository, complete trajectories, submitted patches, and test outputs. Comparing different implementations of the same task helps the grader identify the strongest solutions and distinguish necessary changes from unnecessary complexity. This groupwise comparison also exposes ineffective problem-solving strategies that are difficult to recognize when evaluating candidates individually. Failed attempts provide additional context by revealing unsuccessful strategies, missed requirements, and failure modes, but only passing candidates receive quality rankings.

Both \flash{} and \pro{} experiments use the same pre-RL SFT checkpoint of MiMo-V2.6-Pro as the online grader (Section~\ref{sec:exp_setup}). Our initial implementation used Claude Opus 5, with an average end-to-end grading time of approximately 2{,}000~s per group. Since grading begins only after all rollouts in a group have completed, this added substantial latency to training. Our SFT-trained grader reduces average grading time to approximately 600~s while maintaining good accuracy. We further combine grading with partial-rollout scheduling so that grading completed groups can overlap with rollout generation for other tasks.

\paragraph{Agentic Evidence Gathering.}
Assessing a rollout group requires evidence scattered across long trajectories, patches, and repository files, which is difficult to review in a single model input. Our agentic grader instead gathers evidence iteratively. The grader first reviews a turn-by-turn summary of the rollouts to identify which parts require closer inspection. It then reads relevant portions of individual trajectories and cross-checks submitted patches against repository code and test logs. This allows it to trace how solutions were developed and investigate whether their changes are justified, rather than relying only on a fixed summary. It can also run targeted checks when inspection alone leaves a question unresolved. Negative assessments must cite supporting patch locations, trajectory events, or execution results.

\paragraph{Quality Criteria And Ranking.}
We assess five complementary aspects of code quality beyond test success. Approach suitability evaluates the solution strategy; precision and minimality assess whether its implementation is well-targeted and limited to necessary changes; side effects and codebase consistency assess its impact on existing behavior and maintainability. Together, these criteria guide learning toward sound strategies and well-targeted, merge-ready implementations, with the aim of reducing developer review and revision effort.

The grader first checks whether passing solutions rely on leaked or external answers. Confirmed cases receive zero reward and are excluded from quality ranking. For the remaining candidates, it scores these criteria and flags severe process issues. These assessments determine three quality tiers: $\mathcal{T}_1$ for strong implementations without severe process issues or unresolved regressions, $\mathcal{T}_2$ for intermediate candidates, and $\mathcal{T}_3$ for implementations with major quality defects. Within each tier, weighted criterion scores provide an initial ranking, which the grader can refine through evidence-backed comparisons; candidates remain tied when no distinction is justified. Each candidate's tier and within-tier rank are then mapped to a discount factor $f_i\in(0,1]$ on its positive advantage: $\mathcal{T}_1$ candidates retain full or near-full weight, $\mathcal{T}_2$ candidates receive rank-dependent discounts, and $\mathcal{T}_3$ candidates receive the strongest fixed discount. Tied candidates receive identical factors. The grading stage thus produces a quality-based factor for each remaining passing trajectory, together with supporting evidence, as input to the advantage redistribution in Section~\ref{sec:method_redistribution}. Detailed quality criteria, tiering, ranking, and factor-mapping rules are provided in Appendix~\ref{sec:method_rank_weights}.

\subsection{Sum-Preserving Advantage Redistribution}
\label{sec:method_redistribution}

We use the quality factors $f_i$ to redistribute credit among passing trajectories.

\paragraph{From Downweighting To Redistribution.}
Applying the factors alone would give $\widetilde A_i=f_iA_i$ for $i\in\mathcal{P}$ and leave failed-trajectory advantages unchanged. Since the original advantages sum to zero, this removes positive credit without a corresponding change on the negative side:
\begin{equation}
    D=\sum_{i\in\mathcal{P}}(1-f_i)A_i\geq 0,
    \qquad \sum_{i=1}^{n}\widetilde A_i=-D.
    \label{eq:downweighting_deficit}
\end{equation}
This deficit makes the total magnitude of negative advantages exceed the total positive advantage, weakening reinforcement of successful trajectories relative to the penalties on failed ones. In our experiments, quality-based downweighting without redistribution accompanies rapid growth in policy entropy and trajectory length, together with unstable evaluation performance (Section~\ref{sec:exp_redistribution}).

To favor higher-quality solutions while preserving the total positive advantage assigned to passing trajectories, we redistribute the removed credit among them. Let $S_{+}=\sum_{i\in\mathcal{P}}A_i$ be this sum. We apply a common rescaling factor after quality-based downweighting:
\begin{equation}
    \lambda=\frac{S_{+}}{\sum_{j\in\mathcal{P}}f_jA_j},
    \qquad
    A_i^{\star}=\begin{cases}
        \lambda f_iA_i, & i\in\mathcal{P},\\
        A_i, & i\in\mathcal{F}.
    \end{cases}
    \label{eq:quality_redistribution}
\end{equation}
The denominator is positive for every retained group. Rescaling applies to all passing trajectories; the removed credit is reallocated in proportion to their quality-weighted advantages.

\paragraph{Credit Conservation And Relative Preference.}
\Eqref{eq:quality_redistribution} directly gives
\begin{equation}
    \sum_{i\in\mathcal{P}}A_i^{\star}=S_{+},
    \qquad A_i^{\star}=A_i\ \ (i\in\mathcal{F}),
    \qquad \sum_{i=1}^{n}(A_i^{\star}-A_i)=0.
    \label{eq:advantage_conservation}
\end{equation}
Thus, the change is zero-sum over the passing subset, and the group retains zero mean without modifying failed-trajectory advantages. Because all passes initially have the same advantage, the update has the equivalent form
\begin{equation}
    A_i^{\star}
    =(1-\bar R)\frac{f_i}{\bar f_{\mathcal{P}}},
    \qquad
    \bar f_{\mathcal{P}}=\frac{1}{|\mathcal{P}|}\sum_{j\in\mathcal{P}}f_j,
    \quad i\in\mathcal{P}.
    \label{eq:relative_quality_weights}
\end{equation}
A candidate gains credit when its factor exceeds the passing-set average and gives up credit when its factor falls below that average. Because the same rescaling factor is applied to every passing trajectory, the advantage ratios established by downweighting remain unchanged: $A_i^{\star}/A_j^{\star}=\widetilde A_i/\widetilde A_j=f_i/f_j$ for $i,j\in\mathcal{P}$. Thus, restoring the total positive advantage preserves both the ordering and the relative strength of the quality preferences. If the group contains only one passing candidate, or all passing factors are equal, this formulation reduces to the original outcome advantages. Our method redistributes trajectory-level advantages without requiring matching intermediate states across rollouts.

We use the advantage-space formulation because it directly expresses how quality weighting should change trajectory learning weights. Mean centering after reward shaping already guarantees zero-sum advantages, but does not by itself preserve the original total positive advantage, failed-trajectory advantages, or quality-induced weight ratios. Our rescaling enforces these additional constraints by redistributing removed positive credit among passing trajectories rather than discarding it. Under mean-centered estimation, we derive an exactly equivalent reward transformation from this target advantage allocation, rather than introducing a separate reward-shaping rule (Appendix~\ref{sec:method_reward_equivalence}). The same conservation principle could be applied at the token level by redistributing positive credit among tokens without changing its total.

\subsection{Online Training Integration}
\label{sec:method_integration}

Grading runs asynchronously with rollout collection. Before using a grading result, we check its validity, for example by ensuring that no passing trajectory is missing from the ranking. Unusable results fall back to the original outcome advantages. Confirmed reliance on an external or leaked solution resets the affected reward to zero before group statistics are recomputed; this integrity correction is separate from quality-based redistribution. The resulting sequence advantage is broadcast to the model-generated response tokens throughout each trajectory. Appendix~\ref{sec:method_safeguards} specifies the implementation safeguards and distinguishes them from the exact sum-preserving formulation above.

\section{Experiments}
\label{sec:experiments}

We organize our experiments around two questions: (1) Can \ourtool{} improve code agent performance, implementation quality, and problem-solving behavior? (2) What role does sum-preserving redistribution play in training stability? We initialize our industrial-scale experiments from pre-RL SFT checkpoints of MiMo-V2.6-Flash and MiMo-V2.6-Pro. Our primary controlled study uses code-only RL with \flash{}, complemented by large-scale mixed-task RL with both \flash{} and \pro{}.

\subsection{Experimental Setup}
\label{sec:exp_setup}

\paragraph{Models.}
We initialize RL from pre-RL SFT checkpoints of two industrial-scale mixture-of-experts models: MiMo-V2.6-Flash, with 310B total and 15B active parameters, and MiMo-V2.6-Pro, with 1.02T total and 42B active parameters. We use \flash{} and \pro{} as shorthand for the corresponding model families throughout the experiments. Both model families use the same pre-RL SFT checkpoint of MiMo-V2.6-Pro as the online grader.

\paragraph{Training Configuration.}
Our main experiments use code-only RL with \flash{}, a training batch size of 128, 16 rollouts per prompt, and token-mean loss aggregation. We use the mean-centered advantage estimator and mixed-outcome group filtering described in Section~\ref{sec:method_setup}, with grading performed asynchronously during rollout collection. Separately, we integrate \ourtool{} into industrial-scale mixed-task RL runs of both \flash{} and \pro{}. Each run uses 1{,}568 prompts per update, 16 rollouts per prompt. These runs combine coding with other domains, as detailed in Section~\ref{sec:exp_mixed_rl}.

\paragraph{Benchmarks And Evaluation Protocol.}
Our evaluation covers two public benchmarks.
DeepSWE v1.1~\citep{huang2026deepswe} evaluates long-horizon software development.
SWE-bench Pro~\citep{deng2025swepro} measures challenging repository-level issue resolution. On both benchmarks, we evaluate the baseline and \ourtool{} under identical experimental settings and compare shared training steps, reporting pass rates, mean interaction turns, and mean total token length. Each evaluation uses three samples per task, with mean pass rate reported as \texttt{avg@3}.

\paragraph{Baselines.}
The main \flash{} comparison contrasts a binary-outcome baseline without online quality grading against \ourtool{}, which adds groupwise agentic grading and sum-preserving advantage redistribution. To examine credit balance, we also compare against quality-based downweighting without redistribution or subsequent group centering in Section~\ref{sec:exp_redistribution}. We include Kimi K3, GPT-5.6 Sol, and Claude Opus 5 as external baseline models for the industrial-scale mixed-task RL experiments in Section~\ref{sec:exp_mixed_rl}.

\subsection{Main Experiments With \flash{}}
\label{sec:exp_flash}

\begin{figure}[t]
    \centering
    \includegraphics[width=\linewidth]{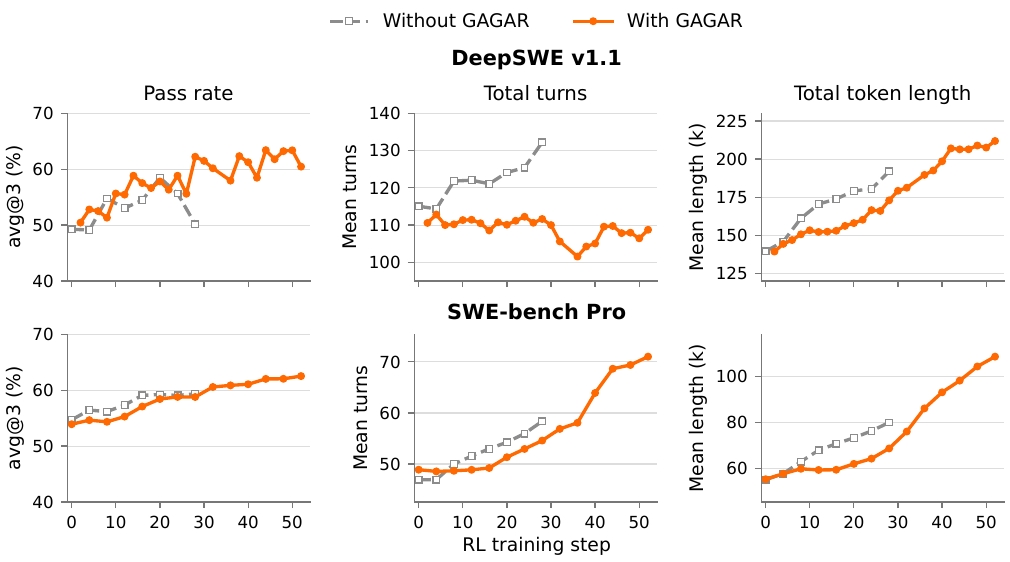}
    \caption{Training dynamics of \flash{} with and without \ourtool{}. Columns show pass rate \texttt{avg@3}, mean main-agent turns, and mean main-agent token length. }
    \label{fig:credit_dynamics}
\end{figure}

Figure~\ref{fig:credit_dynamics} compares task performance and interaction efficiency throughout code-only RL with and without \ourtool{}.

\paragraph{Task Performance.}
We stop the binary-reward baseline at step~28 in response to rapid performance degradation: its DeepSWE pass rate falls from 58.5\% at step~20 to 50.2\%. At step~28, \ourtool{} achieves 62.2\% on DeepSWE v1.1, exceeding the baseline by 12.1 percentage points. With continued training, \ourtool{} reaches a peak DeepSWE pass rate of 63.4\% at step~44. On SWE-bench Pro, the baseline plateaus at approximately 59\% from step~16 through step~28, whereas \ourtool{} continues to improve with further training, reaching 62.5\% at step~52.

\paragraph{Interaction Efficiency And Training Dynamics.}
Without \ourtool{}, DeepSWE trajectory lengths grow rapidly and more trajectories are truncated at the length limit, accompanying the sharp decline in pass rate. With \ourtool{}, DeepSWE turn counts remain roughly stable through step~52 and token length grows more gradually, supporting continued training without the baseline's abrupt deterioration. \ourtool{} uses fewer turns and tokens at every shared DeepSWE checkpoint, with similar reductions on SWE-bench Pro. At step~28, it reduces DeepSWE's mean turn count from 132.3 to 111.6 and mean token length from 191.9k to 172.9k, reductions of 15.6\% and 9.9\%, respectively. On SWE-bench Pro, mean turns decrease from 58.4 to 54.6 and mean token length from 79.9k to 68.6k, reductions of 6.5\% and 14.1\%. The larger pass-rate gains and turn savings on DeepSWE highlight the benefit on difficult, long-horizon tasks requiring many interactions. SWE-bench Pro trajectories are shorter and retain more headroom for growth during continued training.

\subsection{Implementation Quality And Problem-Solving Behavior}
\label{sec:exp_quality}

We evaluate solution quality beyond test success on a fixed random sample of 30 DeepSWE tasks at shared evaluation checkpoints through the end of baseline training. Each task-level group pools up to three rollouts per method, including failures, with anonymized identifiers and randomized order. Claude Opus 5, distinct from the online grader, jointly reviews recorded trajectories, submitted patches, and test outcomes using the five criteria in Section~\ref{sec:method_judging}, scoring and ranking all candidates.

We compute the rubric-weighted quality score defined in Appendix~\ref{sec:method_rank_weights}. We then derive average win rates from the joint groupwise rankings. Finally, we measure how often each method ranks first in its group, splitting cross-method ties equally.

\begin{figure}[htbp]
    \centering
    \includegraphics[width=\linewidth]{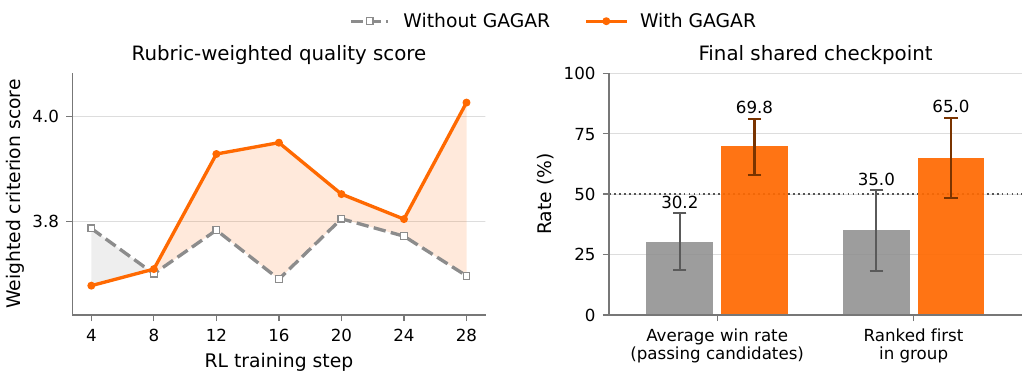}
    \caption{Implementation quality on DeepSWE v1.1. Left: mean rubric-weighted quality score. Right: average win rate among passing candidates and group-first share at the final shared checkpoint.}
    \label{fig:quality_audit}
\end{figure}

Figure~\ref{fig:quality_audit} shows that, at the final shared checkpoint, \ourtool{} achieves a mean quality score of 4.03 versus 3.70 for the baseline, with an average win rate of 69.8\% among passing candidates. It ranks first in 65.0\% of groups after splitting ties. Across the audited checkpoints, improvements are most apparent in implementation precision, minimality of changes, and avoidance of unintended side effects; the fraction of candidates assigned to the highest-quality tier, $\mathcal{T}_1$, increases from 25.4\% to 34.3\%. These assessments support our goal of producing precise, task-scoped, merge-ready implementations that reduce developer review and revision effort.

\subsection{Ablation Of Sum-Preserving Redistribution}
\label{sec:exp_redistribution}

\begin{figure}[t]
    \centering
    \includegraphics[width=\linewidth]{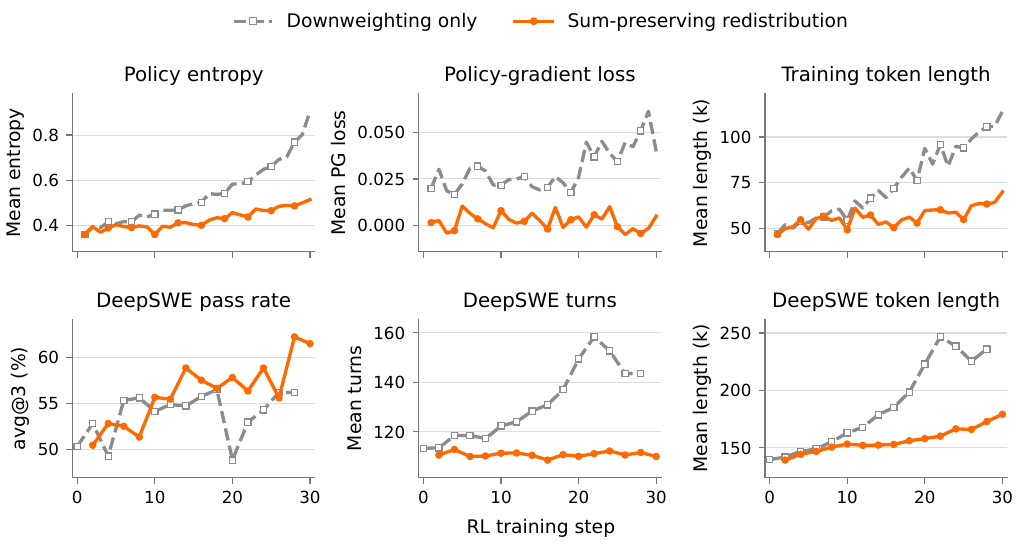}
    \caption{Credit balance and training dynamics in \flash{} RL.}
    \label{fig:redistribution_dynamics}
\end{figure}

We compare code-only \flash{} training with the full redistribution method against a downweighting-only run, which applies quality factors to passing trajectories without restoring the removed positive credit or re-centering the resulting advantages. Figure~\ref{fig:redistribution_dynamics} shows training and evaluation dynamics over the first 30 steps.

Downweighting alone removes positive credit while leaving negative advantages unchanged, shifting their balance toward negative-advantage updates, as described by \eqref{eq:downweighting_deficit}. The downweighting-only run has a consistently higher policy-gradient loss, averaging 0.0304 over steps~1--30, compared with 0.0021 for the full method. This is consistent with the signed nature of the objective: near unit importance ratios, the policy-gradient loss is approximately the negative weighted mean advantage, so an excess of negative advantage can raise its value.

Without redistribution, policy entropy rises from 0.359 at step~1 to 0.905 at step~30, while mean training-rollout length grows from 47.1k to 114.1k tokens. With redistribution, entropy increases more gradually, from 0.358 to 0.513, and mean length grows from 46.5k to 69.9k tokens. The sharper increases without redistribution accompany unstable downstream performance rather than sustained evaluation gains.

On DeepSWE, the downweighting-only pass rate drops from 56.5\% at step~18 to 48.8\% at step~20, then partially recovers to 56.2\% at step~28, compared with 62.2\% for the full method. Its mean interaction count and token length peak at 158.3 turns and 246.7k tokens at step~22. At step~28, these remain elevated at 143.5 turns and 236.0k tokens, versus 111.6 turns and 172.9k tokens with redistribution. Together, these observations support restoring positive credit rather than discarding it when introducing quality preferences.

\subsection{Industrial-Scale Mixed-Task RL}
\label{sec:exp_mixed_rl}

To assess applicability in industrial-scale training, we integrate \ourtool{} into the mixed-task RL training of \mimoflash{} and \mimopro{}, spanning 310B and 1.02T total parameters, respectively. Each update uses 1{,}568 prompts with 16 rollouts per prompt. Grading and redistribution operate on eligible coding-task groups within this heterogeneous workload.

Table~\ref{tab:mixed_rl_results} reports the final \mimoflash{} and \mimopro{} checkpoints from these industrial-scale mixed-task RL runs, alongside K3, GPT-5.6 Sol, and Claude Opus 5. On DeepSWE v1.1, \mimopro{} outperforms Kimi K3 despite having substantially fewer total parameters: 1.02T versus 2.8T~\citep{kimi2026k3}.
It also surpasses GPT-5.6 Sol on SWE-bench Pro, scoring 62.7\% versus 60.5\%. On DeepSWE v1.1, its performance approaches GPT-5.6 Sol and Claude Opus 5, which score 73.0\% and 74.0\%, respectively. These comparisons demonstrate competitive code agent performance relative to frontier models.

\begin{table}[htbp]
    \centering
    \small
    \caption{Coding performance (\%) of \mimoflash{} and \mimopro{} after industrial-scale mixed-task RL with \ourtool{}, compared with external baselines.}
    \label{tab:mixed_rl_results}
    \vspace{4pt}
    \setlength{\tabcolsep}{6pt}
    \renewcommand{\arraystretch}{1.12}
    \begin{tabularx}{\linewidth}{>{\raggedright\arraybackslash}Xrr}
        \toprule
        \rowcolor{tableheaderbg}
        \tableheader{Model} & \tableheader{DeepSWE} & \tableheader{SWE-bench Pro} \\
        \midrule
        K3 & 69.0 & 63.3 \\
        GPT-5.6 Sol & 73.0 & 60.5 \\
        Claude Opus 5 & 74.0 & 79.9 \\
        \addlinespace[3pt]
        \textbf{\mimoflash{}} & 67.9 & 60.9 \\
        \textbf{\mimopro{}} & 71.9 & 62.7 \\
        \bottomrule
    \end{tabularx}
\end{table}

\section{Conclusion}
\label{sec:conclusion}
We presented \ourtool{}, which combines groupwise agentic grading with sum-preserving advantage redistribution to reinforce higher-quality test-passing coding trajectories. Experiments with the 310B-parameter \flash{} model show improved coding performance with fewer interaction turns and lower token usage, while ablations highlight the importance of credit balance for training stability. Industrial-scale mixed-task RL with \flash{} and the 1.02T-parameter \pro{} model demonstrates the applicability of the approach at both model scales.

\begingroup
\small
\bibliography{references}
\endgroup

\clearpage
\appendix
\section{Method Implementation Details}
\label{sec:method_details}

\subsection{Quality Criteria And Rank-To-Weight Mapping}
\label{sec:method_rank_weights}

\begin{table}[htbp]
    \centering
    \small
    \caption{Five dimensions of groupwise implementation-quality assessment, with representative higher- and lower-quality signals.}
    \label{tab:judge_dimensions}
    \vspace{4pt}
    \setlength{\tabcolsep}{6pt}
    \renewcommand{\arraystretch}{1.15}
    \renewcommand{\tabularxcolumn}[1]{m{#1}}
    \begin{tabularx}{\linewidth}{>{\raggedright\arraybackslash\bfseries}m{0.25\linewidth} >{\raggedright\arraybackslash}X >{\raggedright\arraybackslash}X}
        \toprule
        \rowcolor{tableheaderbg}
        \tableheader{Dimension} & \tableheader{Higher-Quality Signals} & \tableheader{Lower-Quality Signals} \\
        \midrule
        Approach Suitability & Root-cause resolution\newline Appropriate solution strategy & Symptom-level workarounds\newline Fundamentally flawed strategy \\
        \addlinespace[5pt]
        Implementation Precision & Correct change locations\newline Complete behavior coverage & Scattered special cases\newline Redundant fallback logic \\
        \addlinespace[5pt]
        Minimality Of Changes & Necessary implementation edits\newline Task-scoped changes & Unrelated refactoring\newline Redundant helpers or branches \\
        \addlinespace[5pt]
        Unintended Side Effects & Preserved surrounding behavior\newline Stable public interfaces & Unrelated behavior changes\newline New risks in core code paths \\
        \addlinespace[5pt]
        Codebase Consistency & Reuse of existing abstractions\newline Consistent code conventions & Ad hoc mechanisms\newline Inconsistent style or comments \\
        \bottomrule
    \end{tabularx}
\end{table}

The grader assigns integer scores from 1 to 5 to the five criteria in Table~\ref{tab:judge_dimensions}. Denote these scores by $s_i^{\mathrm{app}}$, $s_i^{\mathrm{prec}}$, $s_i^{\mathrm{min}}$, $s_i^{\mathrm{side}}$, and $s_i^{\mathrm{style}}$, respectively. The initial ranking uses
\begin{equation}
    W_i=0.30s_i^{\mathrm{app}}+0.25s_i^{\mathrm{prec}}
        +0.20s_i^{\mathrm{min}}+0.15s_i^{\mathrm{side}}+0.10s_i^{\mathrm{style}}.
    \label{eq:judge_initial_score}
\end{equation}
These scores are produced after joint inspection of the group. The grader can revise the initial ordering, but must explain each revision with task-specific evidence. Rankings must respect quality tiers, and ties are allowed within each tier.

\paragraph{Quality Tiers.}
Tiers are recomputed from the criterion scores and flagged issues in a fixed order. We first assign candidates to $\mathcal{T}_3$ if they exhibit a confirmed unrequested rewrite or test-specific workaround, receive the lowest approach-suitability score, or score at most 2 on both minimality and side-effect avoidance. Among the remaining candidates, those scoring at least 4 on every criterion with no severe process issue or unresolved regression belong to $\mathcal{T}_1$; all others belong to $\mathcal{T}_2$.  Confirmed reliance on leaked or external solutions is handled through reward correction before tier assignment.

Let $k_i$ denote the zero-based rank of a candidate's tied group within its tier, and let $K_2$ be the number of tied groups in $\mathcal{T}_2$. The factor map is
\begin{equation}
    f_i=\begin{cases}
        1, & i\in\mathcal{T}_1,\ k_i=0,\\
        f_{\mathrm{runner}}, & i\in\mathcal{T}_1,\ k_i>0,\\
        f_{\max}-(f_{\max}-f_{\min})\dfrac{k_i}{K_2-1},
            & i\in\mathcal{T}_2,\ K_2>1,\\
        f_{\max}, & i\in\mathcal{T}_2,\ K_2=1,\\
        f_{\mathrm{low}}, & i\in\mathcal{T}_3.
    \end{cases}
    \label{eq:rank_factor_map}
\end{equation}
The \flash{} configuration uses $f_{\mathrm{runner}}=0.9$, $f_{\min}=0.4$, $f_{\max}=0.85$, and $f_{\mathrm{low}}=0.2$. All candidates in a tied group receive the same factor. Factors are determined by tied-group ranks rather than by the number of candidates preceding a trajectory.

\subsection{Implementation Safeguards}
\label{sec:method_safeguards}

The training implementation bounds the common rescaling factor in \eqref{eq:quality_redistribution}. Writing $\lambda_{\mathrm{bnd}}=\min(\lambda,\lambda_{\max})$, it computes
\begin{equation}
    B_i=\begin{cases}
        \lambda_{\mathrm{bnd}}f_iA_i, & i\in\mathcal{P},\\
        A_i, & i\in\mathcal{F},
    \end{cases}
    \qquad
    A_i^{\mathrm{train}}=B_i-\frac{1}{n}\sum_{j=1}^{n}B_j.
    \label{eq:bounded_redistribution}
\end{equation}
The \flash{} configuration sets $\lambda_{\max}=1.5$. When the bound is inactive, the centering term is zero and the update equals $A_i^{\star}$. When it is active, centering restores zero group mean and preserves the ordering of passing advantages, but the original positive-advantage sum and failed-trajectory advantages need not be preserved. The exact conservation identities in Section~\ref{sec:method_redistribution} apply to the sum-preserving formulation, not to this bounded branch.

\paragraph{Existing Reward Postprocessing.}
The training system also supports length-based reward adjustments, separate from grading. If such postprocessing produces nonbinary scalar rewards $r_i$, the implementation uses $a_i=r_i-\bar r$, where $\bar r$ is their valid-group mean. Passing candidates are weighted using $a_i^{+}=\max(a_i,0)$, and the common rescaling factor targets $\sum_{i\in\mathcal{P}}a_i^{+}$ rather than a binary-reward sum:
\begin{equation}
    \lambda_{\mathrm{bnd}}=
        \min\!\left(\frac{\sum_{j\in\mathcal{P}}a_j^{+}}
                          {\sum_{j\in\mathcal{P}}f_ja_j^{+}},\lambda_{\max}\right),
    \qquad
    B_i=\begin{cases}
        \lambda_{\mathrm{bnd}}f_i a_i^{+}, & i\in\mathcal{P},\\
        a_i, & i\in\mathcal{F}.
    \end{cases}
    \label{eq:postprocessed_redistribution}
\end{equation}
Rescaling is skipped when the denominator is zero, and the final valid-group mean is subtracted as in \eqref{eq:bounded_redistribution}. Unlike the binary case, passing candidates can have unequal initial advantages; their final ordering then depends on both those advantages and the quality factors. When $r_i=R_i$, this rule reduces to the binary formulation with the same bound.

Infrastructure-invalid trajectories are masked before group statistics are computed. Groups confirmed to have broken task evaluation are excluded from training. When a reference patch is available, the grader may use it to understand the task, but similarity to that patch is not a quality criterion. The reference is not a sampled trajectory and receives no training advantage. Invalid or unavailable grading results leave the original learning signal in place.

\subsection{Equivalent Reward-Space Formulation}
\label{sec:method_reward_equivalence}

Under the mean-centered advantage estimator in Section~\ref{sec:method_setup}, the sum-preserving update admits an exactly equivalent reward transformation. We keep the same valid rollout group and the passing and failing subsets determined by outcome verification.

\paragraph{Equivalent Rewards.}
We construct the equivalent rewards from the target advantages by setting $R_i'=A_i^{\star}+\bar R$, where $\bar R$ is the original pass rate. Substituting \eqref{eq:relative_quality_weights} for passes and $A_i^{\star}=-\bar R$ for failures gives
\begin{equation}
    R_i'=\begin{cases}
        \bar R+(1-\bar R)\dfrac{f_i}{\bar f_{\mathcal{P}}},
            & i\in\mathcal{P},\\
        0, & i\in\mathcal{F}.
    \end{cases}
    \label{eq:equivalent_shaped_rewards}
\end{equation}
Since $\sum_{i\in\mathcal{P}}f_i/\bar f_{\mathcal{P}}=|\mathcal{P}|$, the transformed rewards satisfy $\sum_{i=1}^{n}R_i'=|\mathcal{P}|$ and therefore $\overline{R'}=\bar R$. Subtracting this unchanged group mean gives
\begin{equation}
    R_i'-\overline{R'}=
    \begin{cases}
        (1-\bar R)\dfrac{f_i}{\bar f_{\mathcal{P}}}, & i\in\mathcal{P},\\
        -\bar R, & i\in\mathcal{F},
    \end{cases}
    =A_i^{\star}.
    \label{eq:reward_advantage_equivalence}
\end{equation}
For fixed rollouts and quality factors, this produces the same policy loss and gradient when all other loss terms, masks, and update settings remain unchanged. The transformed rewards depend on the group and may exceed 1; they are optimization signals, not replacements for the binary outcome labels used to identify passes and select groups. Additional reward clipping or reward-standard-deviation normalization generally breaks the equivalence.

\paragraph{Why Direct Reward Discounting Differs.}
Simply setting $\widehat R_i=f_i$ for passing trajectories and $\widehat R_i=0$ for failures changes the group mean to $\bar R\bar f_{\mathcal{P}}$. The resulting advantages are
\begin{equation}
    \widehat A_i=\begin{cases}
        f_i-\bar R\bar f_{\mathcal{P}}, & i\in\mathcal{P},\\
        -\bar R\bar f_{\mathcal{P}}, & i\in\mathcal{F}.
    \end{cases}
    \label{eq:direct_reward_discounting}
\end{equation}
This changes failed-trajectory advantages, generally alters the advantage ratios among passes, and can assign negative advantages to passing trajectories when $f_i<\bar R\bar f_{\mathcal{P}}$. It is therefore not equivalent to our redistribution. Unlike advantage downweighting without subsequent centering, however, direct reward discounting followed by mean subtraction still yields zero-sum advantages. The instability observed in our downweighting-only ablation should not be attributed to reward shaping in general.

\paragraph{Bounded And Postprocessed Updates.}
The implementation in Appendix~\ref{sec:method_safeguards} also admits a reward-space realization. Let $B_i$ be the intermediate value in \eqref{eq:bounded_redistribution} or \eqref{eq:postprocessed_redistribution}, and let $\bar r$ be the original group mean reward. Setting $r_i'=B_i+\bar r$ and then centering gives $r_i'-\overline{r'}=B_i-\bar B=A_i^{\mathrm{train}}$, where $\bar B=n^{-1}\sum_i B_i$. This reproduces the bounded implementation, including its deviations from exact conservation when the rescaling cap is active.

\end{document}